\documentclass{article}

    \PassOptionsToPackage{numbers, compress}{natbib}

    \usepackage[preprint]{neurips_2019}

\usepackage[utf8]{inputenc} 
\usepackage[T1]{fontenc}    
\usepackage{hyperref}       
\usepackage{url}            
\usepackage{booktabs}       
\usepackage{amsfonts}       
\usepackage{nicefrac}       
\usepackage{microtype}      

\usepackage{amsmath}
\usepackage{graphicx}
\usepackage{dsfont}
\usepackage{booktabs}
\usepackage{multirow}
\usepackage{wrapfig}
\usepackage{float}
\usepackage{subfigure}
\usepackage{caption}
\usepackage{ownstyles}
\usepackage{tabularx}
\usepackage{chngcntr}

\newif\ifcomments

\commentstrue

\ifcomments
\newcommand{\comments}[1]{#1}
\else
\newcommand{\comments}[1]{}
\fi

\title{Hamiltonian Neural Networks}

\author{%
  Sam Greydanus \\
  Google Brain\\
  \texttt{sgrey@google.com} \\
   \And
   Misko Dzamba \\
   PetCube \\
   \texttt{mouse9911@gmail.com} \\
   \And
   Jason Yosinski \\
   Uber AI Labs \\
   \texttt{yosinski@uber.com} \\
}

\begin{document}

\maketitle

\begin{abstract}
  Even though neural networks enjoy widespread use, they still struggle to learn the basic laws of physics. How might we endow them with better inductive biases? In this paper, we draw inspiration from Hamiltonian mechanics to train models that learn and respect exact conservation laws in an unsupervised manner. We evaluate our models on problems where conservation of energy is important, including the two-body problem and pixel observations of a pendulum. Our model trains faster and generalizes better than a regular neural network. An interesting side effect is that our model is perfectly reversible in time.
\end{abstract}

\begin{figure}[h!]
\centering
\includegraphics[width=\textwidth]{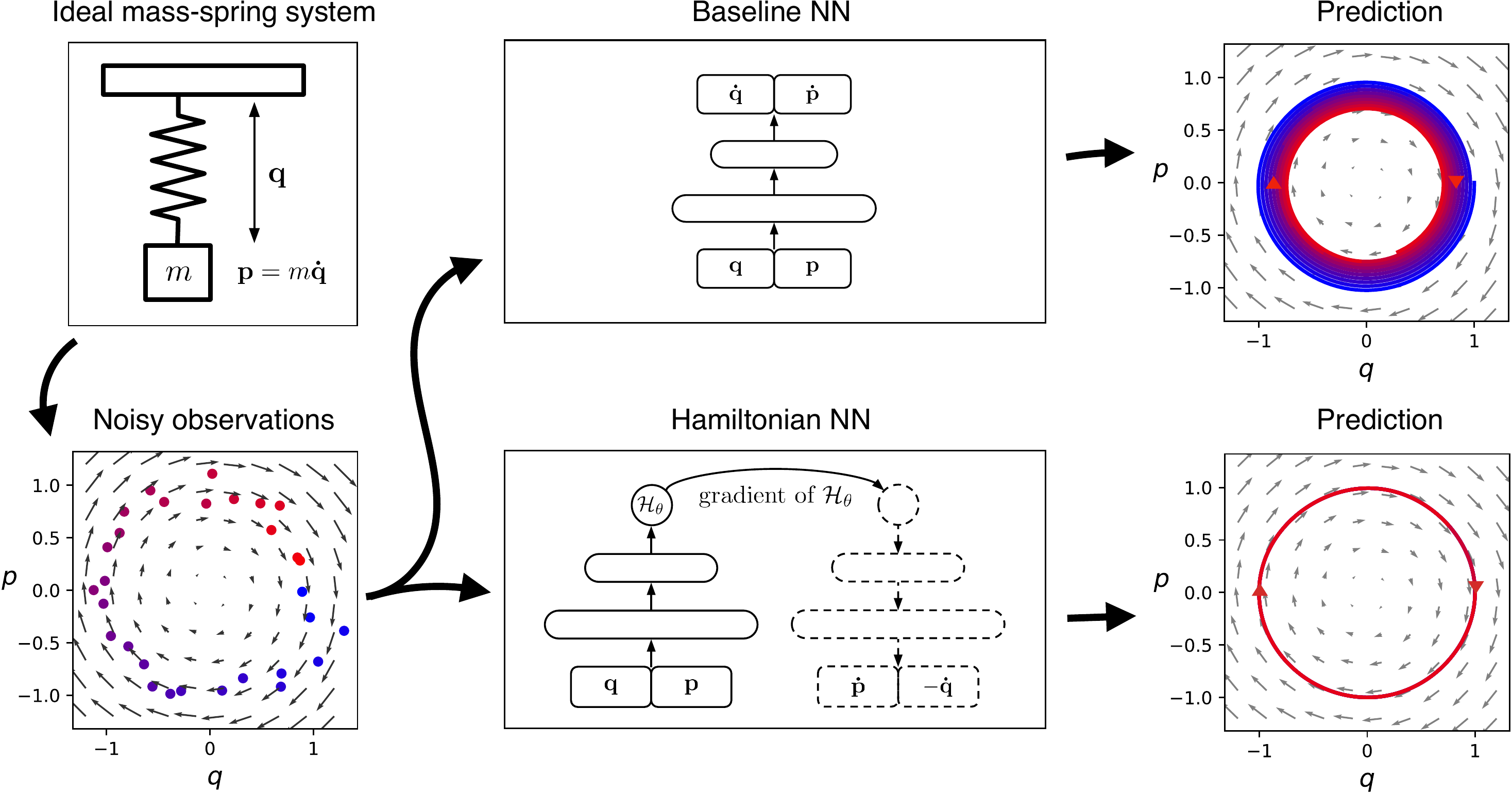}
\caption{
  Learning the Hamiltonian of a mass-spring system. The variables $q$ and $p$ correspond to position and momentum coordinates. As there is no friction, the baseline's inner spiral is due to model errors. By comparison, the Hamiltonian Neural Network learns to \textit{exactly} conserve a quantity that is analogous to total energy.}
\figlabel{fig1}
\end{figure}

\section{Introduction} \seclabel{intro}


Neural networks have a remarkable ability to learn and generalize from data. This lets them excel at tasks such as image classification \cite{krizhevsky2012imagenet-classification-with-deep}, reinforcement learning \cite{yosinski2011evolving-robot-gaits,mnih2013playing-atari-with,silver2017mastering}, and robotic dexterity \cite{Andrychowicz2018Learning, levine2018learning}. Even though these tasks are diverse, they all share the same underlying physical laws. For example, a notion of gravity is important for reasoning about objects in an image, training an RL agent to walk, or directing a robot to manipulate objects. Based on this observation, researchers have become increasingly interested in finding physics priors that transfer across tasks \cite{Watters2017Visual, Santoro2017Simple, Hamrick2018Relational, De2018End, chang2016compositional, tenenbaum-2000-Science-a-global-geometric-framework}.

Untrained neural networks do not have physics priors; they learn approximate physics knowledge directly from data. This generally prevents them from learning \textit{exact} physical laws. Consider the frictionless mass-spring system shown in \figref{fig1}. Here the total energy of the system is being conserved. More specifically, this particular system conserves a quantity proportional to $q^2 + p^2$, where $q$ is the position and $p$ is the momentum of the mass. The baseline neural network in \figref{fig1} learns an approximation of this conservation law, and yet the approximation is imperfect enough that a forward simulation of the system drifts over time to higher or lower energy states.
Can we define a class of neural networks that will precisely conserve energy-like quantities over time?

In this paper, we draw inspiration from Hamiltonian mechanics, a branch of physics concerned with conservation laws and invariances, to define \emph{Hamiltonian Neural Networks}, or \emph{HNNs}. We begin with an equation called the Hamiltonian, which relates the state of a system to some conserved quantity (usually energy) and lets us simulate how the system changes with time. Physicists generally use domain-specific knowledge to find this equation, but here we try a different approach:
\begin{quote}\textit{Instead of crafting the Hamiltonian by hand, we propose parameterizing it with a neural network and then learning it directly from data.}
\end{quote}
Since almost all physical laws can be expressed as conservation laws, our approach is quite general \cite{noether1971invariant}. In practice, our model trains quickly and generalizes well\footnote{We make our code available at \texttt{github.com/greydanus/hamiltonian-nn}.}. \figref{fig1}, for example, shows the outcome of training an HNN on the same mass-spring system. Unlike the baseline model, it learns to conserve an energy-like quantity.

\section{Theory} \seclabel{theory}

\textbf{Predicting dynamics.} The hallmark of a good physics model is its ability to predict changes in a system over time. This is the challenge we now turn to. In particular, our goal is to learn the dynamics of a system using a neural network. The simplest way of doing this is by predicting the next state of a system given the current one. A variety of previous works have taken this path and produced excellent results \cite{tompson2017accelerating, grzeszczuk2000neuroanimator, Watters2017Visual, Santoro2017Simple, Hamrick2018Relational, chang2016compositional}. There are, however, a few problems with this approach.

The first problem is its notion of discrete ``time steps'' that connect neighboring states. Since time is actually continuous, a better approach would be to express dynamics as a set of differential equations and then integrate them from an initial state at $t_0$ to a final state at $t_1$. \eqnref{eqn1} shows how this might be done, letting $\mathbf{S}$ denote the time derivatives of the coordinates of the system\footnote{Any coordinates that describe the state of the system. Later we will use position and momentum $(\textbf{p},\textbf{q})$.}. This approach has been under-explored so far, but techniques like Neural ODEs take a step in the right direction \cite{Chen2018NeuralEquations}.
\begin{equation}
(\mathbf{q}_1,\mathbf{p}_1) ~=~ (\mathbf{q}_0,\mathbf{p}_0) ~+~ \int_{t_0}^{t_1} \mathbf{S}(\mathbf{q},\mathbf{p}) ~~ dt
\eqnlabel{eqn1}
\end{equation}
The second problem with existing methods is that they tend not to learn exact conservation laws or invariant quantities. This often causes them to drift away from the true dynamics of the system as small errors accumulate. The HNN model that we propose ameliorates both of these problems. To see how it does this --- and to situate our work in the proper context ---
we first briefly review Hamiltonian mechanics.

\textbf{Hamiltonian Mechanics.} William Hamilton introduced Hamiltonian mechanics in the 19$^{\textrm{th}}$ century as a mathematical reformulation of classical mechanics. Its original purpose was to express classical mechanics in a more unified and general manner. Over time, though, scientists have applied it to nearly every area of physics from thermodynamics to quantum field theory \cite{reichl1999modern, sakurai1995modern, taylor2005classical}.

In Hamiltonian mechanics, we begin with a set of coordinates $(\mathbf{q},\mathbf{p})$. Usually, $\mathbf{q}=(q_1,...,q_N)$ represents the positions of a set of objects whereas $\mathbf{p}=(p_1,..., p_N)$ denotes their momentum. Note how this gives us $N$ coordinate pairs $(q_1,p_1)...(q_N,p_N)$. Taken together, they offer a complete description of the system. Next, we define a scalar function, $\mathcal{H}(\mathbf{q},\mathbf{p})$ called the Hamiltonian so that
\begin{equation}
\frac{d\mathbf{q}}{dt} ~=~ \frac{\partial \mathcal{H}}{\partial \mathbf{p}}, \quad \frac{d\mathbf{p}}{dt} ~= - \frac{\partial \mathcal{H}}{\partial \mathbf{q}}~.
\eqnlabel{eqn2}
\end{equation}

\eqnref{eqn2} tells us that moving coordinates in the direction $\mathbf{S_{\mathcal{H}}} = \big(\frac{\partial \mathcal{H}}{\partial \mathbf{p}}, -\frac{\partial \mathcal{H}}{\partial \mathbf{q}} \big)$ gives us the time evolution of the system. We can think of $\mathbf{S}$ as a vector field over the inputs of $\mathcal{H}$. In fact, it is a special kind of vector field called a ``symplectic gradient''. Whereas moving in the direction of the gradient of $\mathcal{H}$ changes the output as quickly as possible, moving in the direction of the symplectic gradient \textit{keeps the output exactly constant}. Hamilton used this mathematical framework to relate the position and momentum vectors $(\mathbf{q},\mathbf{p})$ of a system to its total energy $E_{tot}=\mathcal{H}(\mathbf{q},\mathbf{p})$. Then, he found $\mathbf{S_{\mathcal{H}}}$ using \eqnref{eqn2} and obtained the dynamics of the system by integrating this field according to \eqnref{eqn1}. This is a powerful approach because it works for almost any system where the total energy is conserved.

Hamiltonian mechanics, like Newtonian mechanics, can predict the motion of a mass-spring system or a single pendulum. But its true strengths only become apparent when we tackle systems with many degrees of freedom. Celestial mechanics, which are chaotic for more than two bodies, are a good example. A few other examples include many-body quantum systems, fluid simulations, and condensed matter physics \cite{reichl1999modern, sakurai1995modern, taylor2005classical, salmon1988hamiltonian, cohen1997photons, girvin2019modern}.

\textbf{Hamiltonian Neural Networks.} In this paper, we propose learning a parametric function for $\mathcal{H}$ instead of $\mathbf{S_{\mathcal{H}}}$. In doing so, we endow our model with the ability to learn \textit{exactly} conserved quantities from data in an unsupervised manner. During the forward pass, it consumes a set of coordinates and outputs a single scalar ``energy-like'' value. Then, before computing the loss, we take an in-graph gradient of the output with respect to the input coordinates (\figref{appendix:fig6}). It is with respect to this gradient that we compute and optimize an $L_2$ loss (\eqnref{eqn3}).
\begin{equation}
\mathcal{L}_{HNN} =  \bigg \Vert \frac{\partial \mathcal{H_{\theta}}}{ \partial \mathbf{p}} - \frac{\partial \mathbf{q}}{\partial t}  \bigg \Vert _2 +  \bigg \Vert \frac{\partial \mathcal{H_{\theta}}}{ \partial \mathbf{q}} + \frac{\partial \mathbf{p}}{\partial t}  \bigg \Vert_2
\eqnlabel{eqn3}
\end{equation}
For a visual comparison between this approach and the baseline, refer to \figref{fig1} or \figref{appendix:fig6b}. This training procedure allows HNNs to learn conserved quantities analogous to total energy straight from data. Apart from conservation laws, HNNs have several other interesting and potentially useful properties. First, they are perfectly reversible in that the mapping from $(\mathbf{q},\mathbf{p})$ at one time to $(\mathbf{q},\mathbf{p})$ at another time is bijective. Second, we can manipulate the HNN-conserved quantity (analogous to total energy) by integrating along the gradient of $\mathcal{H}$, giving us an interesting counterfactual tool (e.g. ``What would happen if we added 1 Joule of energy?''). We'll discuss these properties later in \secref{useful}.


\section{Learning a Hamiltonian from Data} \seclabel{learning}

Optimizing the gradients of a neural network is a rare approach. There are a few previous works which do this \cite{Wang2018Machine, Schmidt2009Distilling, pukrittayakamee2009simultaneous}, but their scope and implementation details diverge from this work and from one another. With this in mind, our first step was to investigate the empirical properties of HNNs on three simple physics tasks.

\textbf{Task 1: Ideal Mass-Spring.} Our first task was to model the dynamics of the frictionless mass-spring system shown in \figref{fig1}. The system's Hamiltonian is given in \eqnref{eqn4} where $k$ is the spring constant and $m$ is the mass constant. For simplicity, we set $k=m=1$. Then we sampled initial coordinates with total energies uniformly distributed between $[0.2,1]$. We constructed training and test sets of 25 trajectories each and added Gaussian noise with standard deviation $\sigma^2=0.1$ to every data point. Each trajectory had 30 observations; each observation was a concatenation of $(\textbf{q},\textbf{p})$.
\begin{equation}
\mathcal{H} = \frac{1}{2}kq^2 + \frac{p^2}{2m}
\eqnlabel{eqn4}
\end{equation}
\textbf{Task 2: Ideal Pendulum.} Our second task was to model a frictionless pendulum. Pendulums are nonlinear oscillators so they present a slightly more difficult problem. Writing the gravitational constant as $g$ and the length of the pendulum as $l$, the general Hamiltonian is
\begin{equation}
\mathcal{H} = 2 m g l (1-\cos q) + \frac{l^2p^2}{2m}
\eqnlabel{eqn5}
\end{equation}
Once again we set $m=l=1$ for simplicity. This time, we set $g=3$ and sampled initial coordinates with total energies in the range $[1.3,2.3]$. We chose these numbers in order to situate the dataset along the system's transition from linear to nonlinear dynamics. As with Task 1, we constructed training and test sets of 25 trajectories each and added the same amount of noise.

\textbf{Task 3: Real Pendulum.} Our third task featured the position and momentum readings from a real pendulum. We used data from a \textit{Science} paper by \citet{Schmidt2009Distilling} which also tackled the problem of learning conservation laws from data. This dataset was noisier than the synthetic ones and it did not \textit{strictly} obey any conservation laws since the real pendulum had a small amount of friction. Our goal here was to examine how HNNs fared on noisy and biased real-world data.

\subsection{Methods} In all three tasks, we trained our models with a learning rate of $10^{-3}$ and used the Adam optimizer \cite{Kingma2014Adam}. Since the training sets were small, we set the batch size to be the total number of examples. On each dataset we trained two fully-connected neural networks: the first was a baseline model that, given a vector input $(\mathbf{q},\mathbf{p})$ output the vector $(\partial \mathbf{q}/{\partial t}, \partial\mathbf{p}/\partial t)$ directly. The second was an HNN that estimated the same vector using the derivative of a scalar quantity as shown in \eqnref{eqn2} (also see \figref{appendix:fig6}).
Where possible, we used analytic time derivatives as the targets. Otherwise, we calculated finite difference approximations. All of our models had three layers, 200 hidden units, and \texttt{tanh} activations. We trained them for $2000$ gradient steps and evaluated them on the test set.

We logged three metrics: $L_2$ train loss, $L_2$ test loss, and mean squared error (MSE) between the true and predicted total energies. To determine the energy metric, we integrated our models according to \eqnref{eqn1} starting from a random test point. Then we used MSE to measure how much a given model's dynamics diverged from the ground truth. Intuitively, the loss metrics measure our model's ability to fit individual data points while the energy metric measures its stability and conservation of energy over long timespans. To obtain dynamics, we integrated our models with the fourth-order Runge-Kutta integrator in \texttt{scipy.integrate.solve\_ivp} and set the error tolerance to $10^{-9}$ \cite{runge-1895-uber-die-numerische-auflosung}.

\subsection{Results}

\begin{figure}[h!]
\centering
\includegraphics[width=\textwidth]{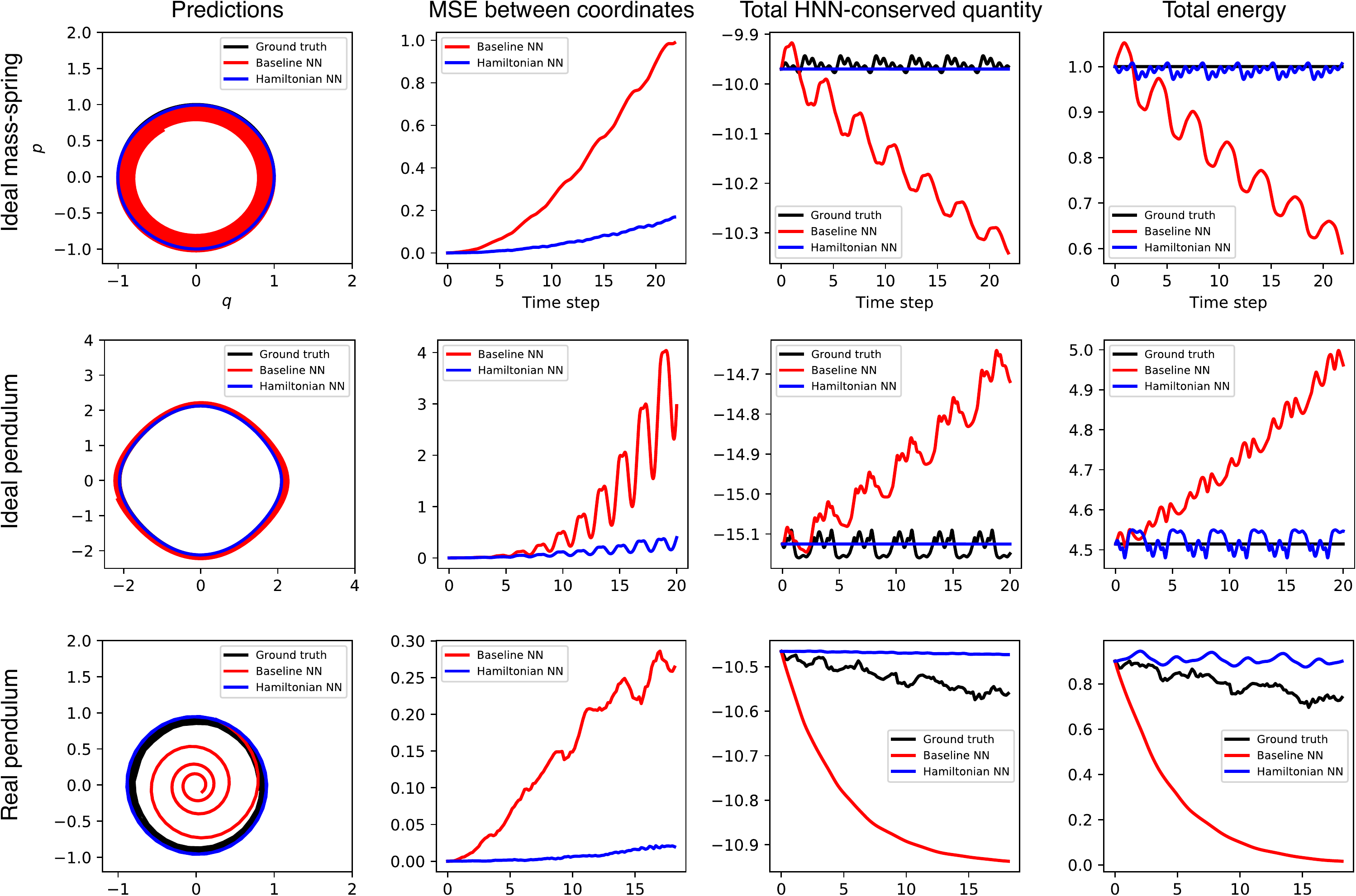}
\caption{
  Analysis of models trained on three simple physics tasks. In the first column, we observe that the baseline model's dynamics gradually drift away from the ground truth. The HNN retains a high degree of accuracy, even obscuring the black baseline in the first two plots. In the second column, the baseline's coordinate MSE error rapidly diverges whereas the HNN's does not. In the third column, we plot the quantity conserved by the HNN. Notice that it closely resembles the total energy of the system, which we plot in the fourth column. In consequence, the HNN roughly conserves total energy whereas the baseline does not.}
\figlabel{fig2}
\end{figure}

We found that HNNs train as quickly as baseline models and converge to similar final losses. \tabref{tab1} shows their relative performance over the three tasks. But even as HNNs tied with the baseline on on loss, they dramatically outperformed it on the MSE energy metric. \figref{fig2} shows why this is the case: as we integrate the two models over time, various errors accumulate in the baseline and it eventually diverges. Meanwhile, the HNN conserves a quantity that closely resembles total energy and diverges more slowly or not at all.

It's worth noting that the quantity conserved by the HNN is not equivalent to the total energy; rather, it's something very close to the total energy. The third and fourth columns of \figref{fig2} provide a useful comparison between the HNN-conserved quantity and the total energy. Looking closely at the spacing of the $y$ axes, one can see that the HNN-conserved quantity has the same scale as total energy, but differs by a constant factor. Since energy is a relative quantity, this is perfectly acceptable\footnote{To see why energy is relative, imagine a cat that is at an elevation of $0$ m in one reference frame and $1$ m in another. Its potential energy (and total energy) will differ by a constant factor depending on frame of reference.}.

The total energy plot for the real pendulum shows another interesting pattern. Whereas the ground truth data does not quite conserve total energy, the HNN roughly conserves this quantity. This, in fact, is a fundamental limitation of HNNs: they assume a conserved quantity exists and thus are unable to account for things that violate this assumpation, such as friction. In order to account for friction, we would need to model it separately from the HNN.

\section{Modeling Larger Systems}  \seclabel{larger}

Having established baselines on a few simple tasks, our next step was to tackle a larger system involving more than one pair of $(p,q)$ coordinates. One well-studied problem that fits this description is the two-body problem, which requires four $(p,q)$ pairs.
\begin{equation}
\mathcal{H} = \frac{|\mathbf{p_{CM}}|^2}{m_1 + m_2} + \frac{|\mathbf{p_1}|^2 + |\mathbf{p_2}|^2}{2\mu} + g\frac{m_1m_2}{|\mathbf{q_1}-\mathbf{q_2}|^2}
\eqnlabel{eqn6}
\end{equation}
\textbf{Task 4: Two-body problem.} In the two-body problem, point particles interact with one another via an attractive force such as gravity. Once again, we let $g$ be the gravitational constant and $m$ represent mass. \eqnref{eqn6} gives the Hamiltonian of the system where $\mu$ is the reduced mass and $\mathbf{p_{CM}}$ is the momentum of the center of mass. As in previous tasks, we set $m_1=m_2=g=1$ for simplicity. Furthermore, we restricted our experiments to systems where the momentum of the center of mass was zero. Even so, with eight degrees of freedom (given by the $x$ and $y$ position and momentum coordinates of the two bodies) this system represented an interesting challenge.

\subsection{Methods} \seclabel{orbit-methods}
Our first step was to generate a dataset of 1000 near-circular, two-body trajectories. We initialized every trajectory with center of mass zero, total momentum zero, and radius $r=\Vert \mathbf{q_2}-\mathbf{q_1} \Vert$ in the range $[0.5,1.5]$. In order to control the level of numerical stability, we chose initial velocities that gave perfectly circular orbits and then added Gaussian noise to them. We found that scaling this noise by a factor of $\sigma^2=0.05$ produced trajectories with a good balance between stability and diversity.

We used fourth-order Runge-Kutta integration to find 200 trajectories of 50 observations each and then performed an 80/20\% train/test set split over trajectories. Our models and training procedure were identical to those described in \secref{learning} except this time we trained for 10,000 gradient steps and used a batch size of 200.

\subsection{Results}

\begin{wrapfigure}[18]{R}{0.56\textwidth}
\centering
\vspace{-.3cm}%
\begin{tabular}{c}
\setlength{\tabcolsep}{0pt}
\includegraphics[width=.55\textwidth]{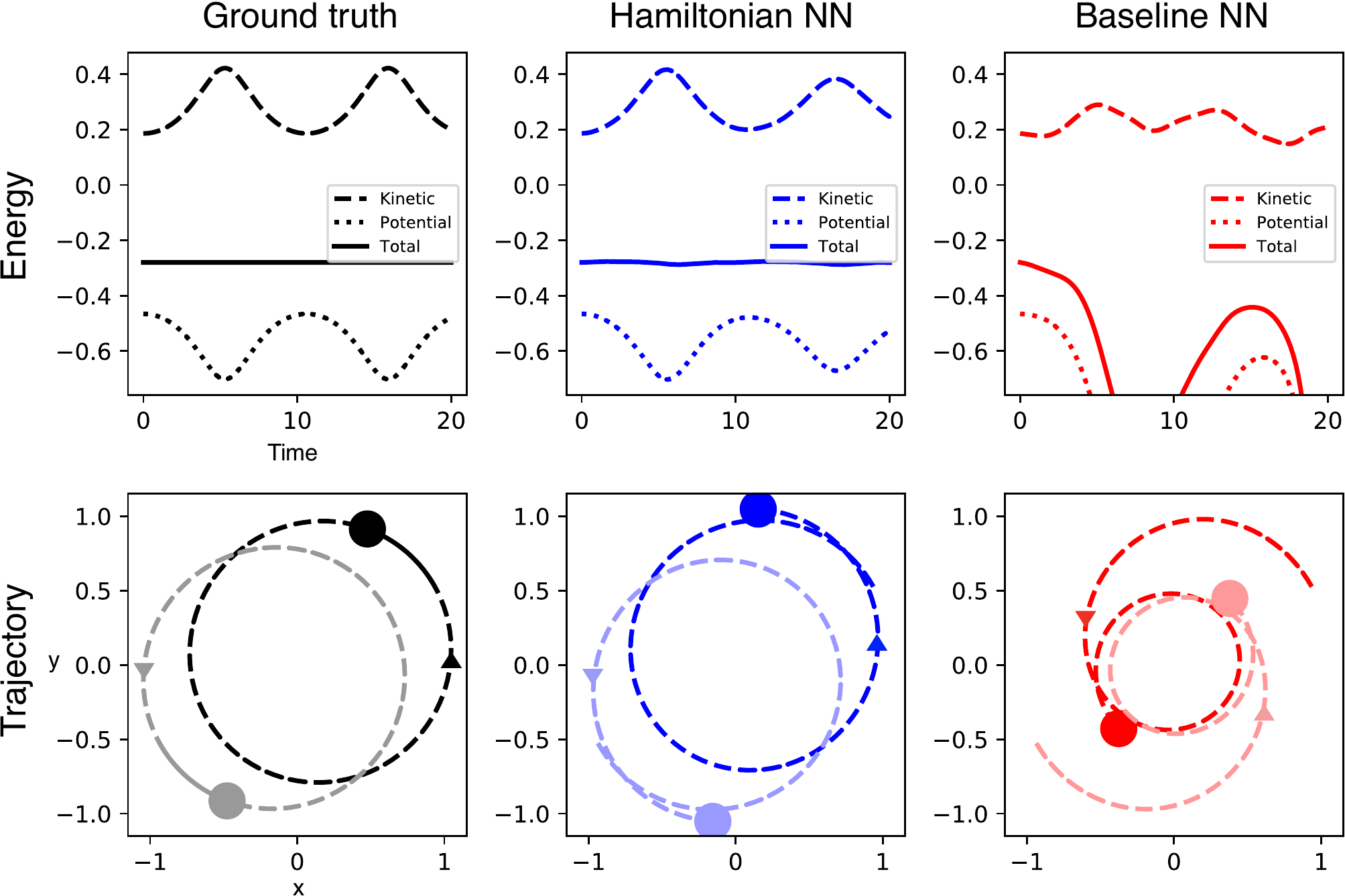}
\end{tabular}
\caption{
    Analysis of an example 2-body trajectory. The dynamics of the baseline model do not conserve total energy and quickly diverge from ground truth. The HNN, meanwhile, approximately conserves total energy and accrues a small amount of error after one full orbit.
}
\figlabel{fig3}
\end{wrapfigure}

The HNN model scaled well to this system. The first row of \figref{fig3} suggests that it learned to conserve a quantity nearly equal to the total energy of the system whereas the baseline model did not.

The second row of \figref{fig3} gives a qualitative comparison of trajectories. After one orbit, the baseline dynamics have completely diverged from the ground truth whereas the HNN dynamics have only accumulated a small amount of error. As we continue to integrate up to $t=50$ and beyond (\figref{appendix:fig8}), both models diverge but the HNN does so at a much slower rate. Even as the HNN diverges from the ground truth orbit, its total energy remains stable rather than decaying to zero or spiraling to infinity. We report quantitative results for this task in \tabref{tab1}. Both train and test losses of the HNN model were about an order of magnitude lower than those of the baseline. The HNN did a better job of conserving total energy, with an energy MSE that was several orders of magnitude below the baseline.

Having achieved success on the two-body problem, we ran the same set of experiments on the chaotic three-body problem. We show preliminary results in Appendix \ref{appendix:orbits} where once again the HNN outperforms its baseline by a considerable margin. We opted to focus on the two-body results here because the three-body results still need improvement.

\section{Learning a Hamiltonian from Pixels}   \seclabel{pixels}

One of the key strengths of neural networks is that they can learn abstract representations directly from high-dimensional data such as pixels or words. Having trained HNN models on position and momentum coordinates, we were eager to see whether we could train them on arbitrary coordinates like the latent vectors of an autoencoder.

\textbf{Task 5: Pixel Pendulum.} With this in mind, we constructed a dataset of pixel observations of a pendulum and then combined an autoencoder with an HNN to model its dynamics. To our knowledge this is the first instance of a Hamiltonian learned directly from pixel data.

\subsection{Methods}

In recent years, OpenAI Gym has been widely adopted by the machine learning community as a means for training and evaluating reinforcement learning agents \cite{Brockman2016OpenAI}. Some works have even trained world models on these environments \cite{Ha2018Recurrent,Hafner2018Learning}. Seeing these efforts as related and complimentary to our work, we used OpenAI Gym's \texttt{Pendulum-v0} environment in this experiment. 

First, we generated $200$ trajectories of $100$ frames each\footnote{Choosing the ``no torque'' action at every timestep.}. We required that the maximum absolute displacement of the pendulum arm be $\frac{\pi}{6}$ radians. Starting from $400\textrm{ x }400\textrm{ x }3$ RGB pixel observations, we cropped, desaturated, and downsampled them to $28\textrm{ x }28\textrm{ x }1$ frames and concatenated each frame with its successor so that the input to our model was a tensor of shape $\texttt{batch}\textrm{ x }28\textrm{ x }28\textrm{ x }2$. We used two frames so that velocity would be observable from the input. Without the ability to observe velocity, an autoencoder without recurrence would be unable to ascertain the system's full state space.

In designing the autoencoder portion of the model, our main objective was simplicity and trainability. We chose to use fully-connected layers in lieu of convolutional layers because they are simpler. Furthermore, convolutional layers sometimes struggle to extract even simple position information \cite{Liu2018Intriguing}. Both the encoder and decoder were composed of four fully-connected layers with \texttt{relu} activations and residual connections. We used $200$ hidden units on all layers except the latent vector $\mathbf{z}$, where we used two units. As for the HNN component of this model, we used the same architecture and parameters as described in \secref{learning}. Unless otherwise specified, we used the same training procedure as described in \secref{orbit-methods}. We found that using a small amount of weight decay, $10^{-5}$ in this case, was beneficial.

\textbf{Losses.} The most notable difference between this experiment and the others was the loss function. This loss function was composed of three terms: the first being the HNN loss, the second being a classic autoencoder loss ($L_2$ loss over pixels), and the third being an auxiliary loss on the autoencoder's latent space:
\begin{equation}
\mathcal{L}_{CC} =  \big \Vert \mathbf{z}^{t}_{\mathbf{p}} - (\mathbf{z}^{t}_{\mathbf{q}}-z^{t+1}_{\mathbf{q}})  \big \Vert _2
\eqnlabel{eqn7}
\end{equation}
The purpose of the auxiliary loss term, given in \eqnref{eqn7}, was to make the second half of $\mathbf{z}$, which we'll label $\mathbf{z}_{\mathbf{p}}$, resemble the derivatives of the first half of $\mathbf{z}$, which we'll label $\mathbf{z}_{\mathbf{q}}$. This loss encouraged the latent vector $(\mathbf{z}_{\mathbf{q}}, \mathbf{z}_{\mathbf{p}})$ to have roughly same properties as canonical coordinates $(\mathbf{q}, \mathbf{p})$. These properties, measured by the Poisson bracket relations, are necessary for writing a Hamiltonian. We found that the auxiliary loss did not degrade the autoencoder's performance. Furthermore, it is not domain-specific and can be used with any autoencoder with an even-sized latent space.

\begin{figure}[h!]
\centering
\includegraphics[width=\textwidth]{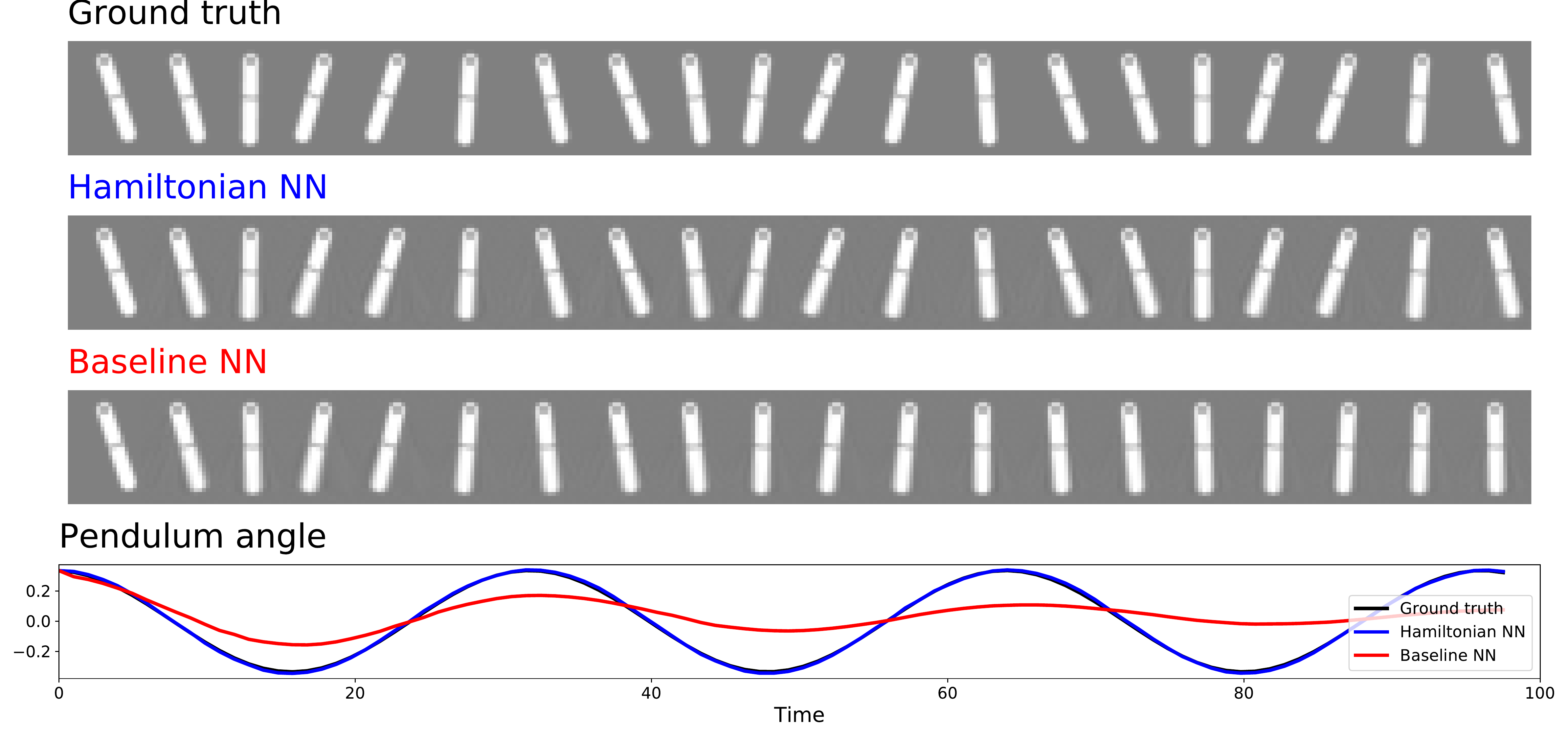}
\caption{Predicting the dynamics of the pixel pendulum. We train an HNN and its baseline to predict dynamics in the latent space of an autoencoder. Then we project to pixel space for visualization. The baseline model rapidly decays to lower energy states whereas the HNN remains close to ground truth even after hundreds of frames. It mostly obscures the ground truth line in the bottom plot.}
\figlabel{fig4}
\end{figure}

\subsection{Results}

Unlike the baseline model, the HNN learned to conserve a scalar quantity analogous to the total energy of the system. This enabled it to predict accurate dynamics for the system over much longer timespans. \figref{fig4} shows a qualitative comparison of trajectories predicted by the two models. As in previous experiments, we computed these dynamics using \eqnref{eqn2} and a fourth-order Runge-Kutta integrator. Unlike previous experiments, we performed this integration in the latent space of the autoencoder. Then, after integration, we projected to pixel space using the decoder network. The HNN and its baseline reached comparable train and test losses, but once again, the HNN dramatically outperformed the baseline on the energy metric (\tabref{tab1}).


\begin{table}[h!]
\centering
\caption{Quantitative results across all five tasks. Whereas the HNN is competitive with the baseline on train/test loss, it dramatically outperforms the baseline on the energy metric. All values are multiplied by $10^{3}$ unless noted otherwise. See Appendix \ref{appendix:learning} for a note on train/test split for Task 3.}
\tablabel{tab1}
\begin{tabular}{@{}lllllll@{}}
\toprule
& Train loss & & Test loss & & Energy& \\ \midrule
Task & Baseline & HNN & Baseline & HNN & Baseline & HNN \\ \midrule
1: Ideal mass-spring & $37\pm2$ & $37\pm2$ & $37\pm2$ & $\mathbf{36\pm2}$ & $170\pm20$ & $\mathbf{.38\pm.1}$  \\
2: Ideal pendulum   & $33\pm2$ & $33\pm2$ & $\mathbf{35\pm2}$   & $36\pm2$ & $42\pm10$ & $\mathbf{25\pm5}$   \\
3: Real pendulum   & $2.7\pm.2$ & $9.2\pm.5$ & $\mathbf{2.2\pm.3}$   & $6.0\pm.6$ & $390\pm7$  & $\mathbf{14\pm5}$  \\
4: Two body ($\times 10^{6}$) & $33\pm1$ & $3.0\pm.1$ & $30\pm.1$ & $\mathbf{2.8\pm.1}$ & $6.3e4\pm3e4$ & $\mathbf{39\pm5}$  \\
5: Pixel pendulum   & $18\pm.2$ & $19\pm.2$ & $\mathbf{17\pm.3}$   & $18\pm.3$ & $9.3\pm1$ & $\mathbf{.15\pm.01}$   \\ \bottomrule
\end{tabular}
\end{table}

\section{Useful properties of HNNs}   \seclabel{useful}

While the main purpose of HNNs is to endow neural networks with better physics priors, in this section we ask what other useful properties these models might have.

\begin{wrapfigure}[18]{R}{0.40\textwidth}
\centering
\vspace{-.7cm}%
\hspace{1cm}%
\begin{tabular}{c}
\setlength{\tabcolsep}{0pt}
\includegraphics[width=0.35\textwidth]{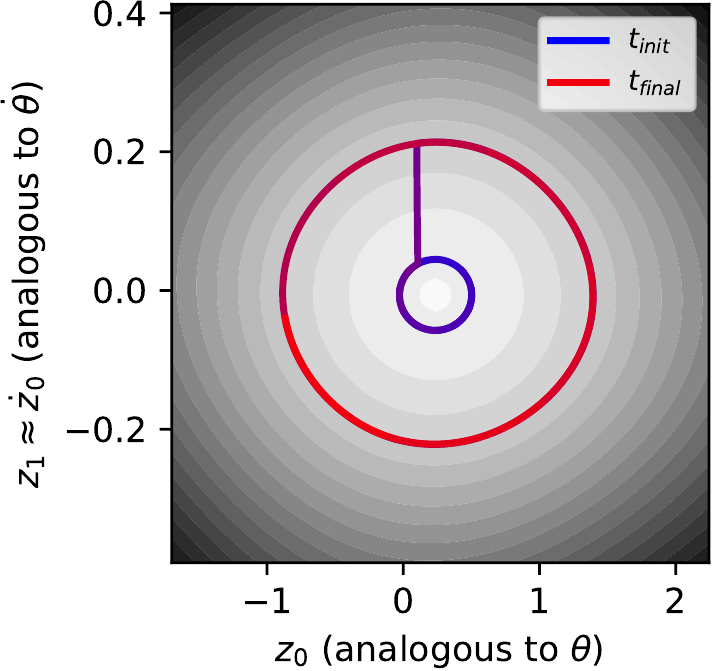}
\end{tabular}
\caption{
  Visualizing integration in the latent space of the Pixel Pendulum model. We alternately integrate $\mathbf{S_{\mathcal{H}}}$ at low energy (blue circle), $\mathbf{R_{\mathcal{H}}}$ (purple line), and then $\mathbf{S_{\mathcal{H}}}$ at higher energy (red circle).
}
\figlabel{fig5}
\end{wrapfigure}

\textbf{Adding and removing energy.} So far, we have seen that integrating the symplectic gradient of the Hamiltonian can give us the time evolution of a system but we have not tried following the Riemann gradient $\mathbf{R_{\mathcal{H}}} = \big(\frac{\partial \mathcal{H}}{\partial \mathbf{q}},  \frac{\partial \mathcal{H}}{\partial \mathbf{p}} \big)$.
Intuitively, this corresponds to adding or removing some of the HNN-conserved quantity from the system. It's especially interesting to alternate between integrating $\mathbf{R_{\mathcal{H}}}$ and $\mathbf{S_{\mathcal{H}}}$. \figref{fig5} shows how we can take advantage of this effect to ``bump'' the pendulum to a higher energy level. We could imagine using this technique to answer counterfactual questions e.g. ``What would have happened if we applied a torque?''

\textbf{Perfect reversibility.} As neural networks have grown in size, the memory consumption of transient activations, the intermediate activations saved for backpropagation, has become a notable bottleneck. Several works propose semi-reversible models that construct one layer's activations from the activations of the next \cite{Gomez2017Reversible,mackay2018reversible, jacobsen2018revnet}. Neural ODEs also have this property \cite{Chen2018NeuralEquations}. Many of these models are only approximately reversible: their mappings are not quite bijective. Unlike those methods, our approach is guaranteed to produce trajectories that are perfectly reversible through time. We can simply refer to a result from Hamiltonian mechanics called Liouville's Theorem: \textit{the density of particles in phase space is constant}. What this implies is that any mapping $(\mathbf{q}_0,\mathbf{p}_0) \rightarrow (\mathbf{q}_1,\mathbf{p}_1)$ is bijective/invertible.

\section{Related work}   \seclabel{related}

\textbf{Learning physical laws from data.} \citet{Schmidt2009Distilling} used a genetic algorithm to search a space of mathematical functions for conservation laws and recovered the Lagrangians and Hamiltonians of several real systems. We were inspired by their approach, but used a neural neural network to avoid constraining our search to a set of hand-picked functions. Two recent works are similar to this paper in that the authors sought to uncover physical laws from data using neural networks \cite{Iten2018Discovering, bondesan2019learning}. Unlike our work, they did not explicitly parameterize Hamiltonians.

\textbf{Physics priors for neural networks.} A wealth of previous works have sought to furnish neural networks with better physics priors. Many of these works are domain-specific: the authors used domain knowledge about molecular dynamics \cite{rupp2012fast,smith2017ani,chmiela2017machine, pukrittayakamee2009simultaneous}, quantum mechanics \cite{schutt2017quantum}, or robotics \cite{lutter2019deep} to help their models train faster or generalize. Others, such as Interaction Networks or Relational Networks were meant to be fully general \cite{Watters2017Visual, Santoro2017Simple, battaglia2016interaction}. Here, we also aimed to keep our approach fully general while introducing a strong and theoretically-motivated prior.

\textbf{Modeling energy surfaces.} Physicists, particularly those studying molecular dynamics, have seen success using neural networks to model energy surfaces \cite{behler2011neural, gastegger2015high, schutt2017quantum, yao2018tensormol}. In particular, several works have shown dramatic computation speedups compared to density functional theory \cite{rupp2012fast,smith2017ani,chmiela2017machine}. Molecular dynamics researchers integrate the derivatives of energy in order to obtain dynamics, just as we did in this work. A key difference between these approaches and our own is that 1) we emphasize the Hamiltonian formalism 2) we optimize the gradients of our model (though some works do optimize the gradients of a molecular dynamics model \cite{Wang2018Machine, pukrittayakamee2009simultaneous}).

\section{Discussion}   \seclabel{discuss}



Whereas Hamiltonian mechanics is an old and well-established theory, the science of deep learning is still in its infancy. Whereas Hamiltonian mechanics describes the real world from first principles, deep learning does so starting from data. We believe that Hamiltonian Neural Networks, and models like them, represent a promising way of bringing together the strengths of both approaches.

\section{Acknowledgements}   \seclabel{discuss}

Sam Greydanus would like to thank the Google AI Residency Program for providing extraordinary mentorship and resources. The authors would like to thank Nic Ford, Trevor Gale, Rapha Gontijo Lopes, Keren Gu, Ben Caine, Mark Woodward, Stephan Hoyer, Jascha Sohl-Dickstein, and many others for insightful conversations and support.

Special thanks to James and Judy Greydanus for their feedback and support from beginning to end.

\bibliography{hnn}

\begin{thebibliography}{45}
\providecommand{\natexlab}[1]{#1}
\providecommand{\url}[1]{\texttt{#1}}
\expandafter\ifx\csname urlstyle\endcsname\relax
  \providecommand{\doi}[1]{doi: #1}\else
  \providecommand{\doi}{doi: \begingroup \urlstyle{rm}\Url}\fi

\bibitem[Andrychowicz et~al.(2018)Andrychowicz, Baker, Chociej, Jozefowicz,
  McGrew, Pachocki, Petron, Plappert, Powell, Ray,
  et~al.]{Andrychowicz2018Learning}
Andrychowicz, M., Baker, B., Chociej, M., Jozefowicz, R., McGrew, B., Pachocki,
  J., Petron, A., Plappert, M., Powell, G., Ray, A., et~al.
\newblock Learning dexterous in-hand manipulation.
\newblock \emph{arXiv preprint arXiv:1808.00177}, 2018.

\bibitem[Battaglia et~al.(2016)Battaglia, Pascanu, Lai, Rezende,
  et~al.]{battaglia2016interaction}
Battaglia, P., Pascanu, R., Lai, M., Rezende, D.~J., et~al.
\newblock Interaction networks for learning about objects, relations and
  physics.
\newblock In \emph{Advances in neural information processing systems}, pp.\
  4502--4510, 2016.

\bibitem[Behler(2011)]{behler2011neural}
Behler, J.
\newblock Neural network potential-energy surfaces in chemistry: a tool for
  large-scale simulations.
\newblock \emph{Physical Chemistry Chemical Physics}, 13\penalty0
  (40):\penalty0 17930--17955, 2011.

\bibitem[Bondesan \& Lamacraft(2019)Bondesan and
  Lamacraft]{bondesan2019learning}
Bondesan, R. and Lamacraft, A.
\newblock Learning symmetries of classical integrable systems.
\newblock \emph{arXiv preprint arXiv:1906.04645}, 2019.

\bibitem[Brockman et~al.(2016)Brockman, Cheung, Pettersson, Schneider,
  Schulman, Tang, and Zaremba]{Brockman2016OpenAI}
Brockman, G., Cheung, V., Pettersson, L., Schneider, J., Schulman, J., Tang,
  J., and Zaremba, W.
\newblock Openai gym.
\newblock \emph{arXiv preprint arXiv:1606.01540}, 2016.

\bibitem[Chang et~al.(2016)Chang, Ullman, Torralba, and
  Tenenbaum]{chang2016compositional}
Chang, M.~B., Ullman, T., Torralba, A., and Tenenbaum, J.~B.
\newblock A compositional object-based approach to learning physical dynamics.
\newblock \emph{arXiv preprint arXiv:1612.00341}, 2016.

\bibitem[Chen et~al.(2018)Chen, Rubanova, Bettencourt, and
  Duvenaud]{Chen2018NeuralEquations}
Chen, T.~Q., Rubanova, Y., Bettencourt, J., and Duvenaud, D.~K.
\newblock Neural ordinary differential equations.
\newblock pp.\  6571--6583, 2018.
\newblock URL
  \url{http://papers.nips.cc/paper/7892-neural-ordinary-differential-equations.pdf}.

\bibitem[Chmiela et~al.(2017)Chmiela, Tkatchenko, Sauceda, Poltavsky,
  Sch{\"u}tt, and M{\"u}ller]{chmiela2017machine}
Chmiela, S., Tkatchenko, A., Sauceda, H.~E., Poltavsky, I., Sch{\"u}tt, K.~T.,
  and M{\"u}ller, K.-R.
\newblock Machine learning of accurate energy-conserving molecular force
  fields.
\newblock \emph{Science advances}, 3\penalty0 (5):\penalty0 e1603015, 2017.

\bibitem[Cohen-Tannoudji et~al.(1997)Cohen-Tannoudji, Dupont-Roc, and
  Grynberg]{cohen1997photons}
Cohen-Tannoudji, C., Dupont-Roc, J., and Grynberg, G.
\newblock Photons and atoms-introduction to quantum electrodynamics.
\newblock \emph{Photons and Atoms-Introduction to Quantum Electrodynamics, by
  Claude Cohen-Tannoudji, Jacques Dupont-Roc, Gilbert Grynberg, pp. 486. ISBN
  0-471-18433-0. Wiley-VCH, February 1997.}, pp.\  486, 1997.

\bibitem[de~Avila Belbute-Peres et~al.(2018)de~Avila Belbute-Peres, Smith,
  Allen, Tenenbaum, and Kolter]{De2018End}
de~Avila Belbute-Peres, F., Smith, K., Allen, K., Tenenbaum, J., and Kolter,
  J.~Z.
\newblock End-to-end differentiable physics for learning and control.
\newblock In \emph{Advances in Neural Information Processing Systems}, pp.\
  7178--7189, 2018.

\bibitem[Gastegger \& Marquetand(2015)Gastegger and
  Marquetand]{gastegger2015high}
Gastegger, M. and Marquetand, P.
\newblock High-dimensional neural network potentials for organic reactions and
  an improved training algorithm.
\newblock \emph{Journal of chemical theory and computation}, 11\penalty0
  (5):\penalty0 2187--2198, 2015.

\bibitem[Girvin \& Yang(2019)Girvin and Yang]{girvin2019modern}
Girvin, S.~M. and Yang, K.
\newblock \emph{Modern condensed matter physics}.
\newblock Cambridge University Press, 2019.

\bibitem[Gomez et~al.(2017)Gomez, Ren, Urtasun, and
  Grosse]{Gomez2017Reversible}
Gomez, A.~N., Ren, M., Urtasun, R., and Grosse, R.~B.
\newblock The reversible residual network: Backpropagation without storing
  activations.
\newblock In \emph{Advances in neural information processing systems}, pp.\
  2214--2224, 2017.

\bibitem[Grzeszczuk()]{grzeszczuk2000neuroanimator}
Grzeszczuk, R.
\newblock \emph{NeuroAnimator: fast neural network emulation and control of
  physics-based models.}
\newblock University of Toronto.

\bibitem[Ha \& Schmidhuber(2018)Ha and Schmidhuber]{Ha2018Recurrent}
Ha, D. and Schmidhuber, J.
\newblock Recurrent world models facilitate policy evolution.
\newblock In \emph{Advances in Neural Information Processing Systems}, pp.\
  2450--2462, 2018.

\bibitem[Hafner et~al.(2018)Hafner, Lillicrap, Fischer, Villegas, Ha, Lee, and
  Davidson]{Hafner2018Learning}
Hafner, D., Lillicrap, T., Fischer, I., Villegas, R., Ha, D., Lee, H., and
  Davidson, J.
\newblock Learning latent dynamics for planning from pixels.
\newblock \emph{arXiv preprint arXiv:1811.04551}, 2018.

\bibitem[Hamrick et~al.(2018)Hamrick, Allen, Bapst, Zhu, McKee, Tenenbaum, and
  Battaglia]{Hamrick2018Relational}
Hamrick, J.~B., Allen, K.~R., Bapst, V., Zhu, T., McKee, K.~R., Tenenbaum,
  J.~B., and Battaglia, P.~W.
\newblock Relational inductive bias for physical construction in humans and
  machines.
\newblock \emph{arXiv preprint arXiv:1806.01203}, 2018.

\bibitem[Iten et~al.(2018)Iten, Metger, Wilming, Del~Rio, and
  Renner]{Iten2018Discovering}
Iten, R., Metger, T., Wilming, H., Del~Rio, L., and Renner, R.
\newblock Discovering physical concepts with neural networks.
\newblock \emph{arXiv preprint arXiv:1807.10300}, 2018.

\bibitem[Jacobsen et~al.(2018)Jacobsen, Smeulders, and
  Oyallon]{jacobsen2018revnet}
Jacobsen, J.-H., Smeulders, A., and Oyallon, E.
\newblock i-revnet: Deep invertible networks.
\newblock \emph{arXiv preprint arXiv:1802.07088}, 2018.

\bibitem[Kingma \& Ba(2014)Kingma and Ba]{Kingma2014Adam}
Kingma, D.~P. and Ba, J.
\newblock Adam: A method for stochastic optimization.
\newblock \emph{International Conference on Learning Representations}, 2014.

\bibitem[Krizhevsky et~al.(2012)Krizhevsky, Sutskever, and
  Hinton]{krizhevsky2012imagenet-classification-with-deep}
Krizhevsky, A., Sutskever, I., and Hinton, G.
\newblock Imagenet classification with deep convolutional neural networks.
\newblock In \emph{Advances in Neural Information Processing Systems 25}, pp.\
  1106--1114, 2012.

\bibitem[Levine et~al.(2018)Levine, Pastor, Krizhevsky, Ibarz, and
  Quillen]{levine2018learning}
Levine, S., Pastor, P., Krizhevsky, A., Ibarz, J., and Quillen, D.
\newblock Learning hand-eye coordination for robotic grasping with deep
  learning and large-scale data collection.
\newblock \emph{The International Journal of Robotics Research}, 37\penalty0
  (4-5):\penalty0 421--436, 2018.

\bibitem[Liu et~al.(2018)Liu, Lehman, Molino, Such, Frank, Sergeev, and
  Yosinski]{Liu2018Intriguing}
Liu, R., Lehman, J., Molino, P., Such, F.~P., Frank, E., Sergeev, A., and
  Yosinski, J.
\newblock An intriguing failing of convolutional neural networks and the
  coordconv solution.
\newblock In \emph{Advances in Neural Information Processing Systems}, pp.\
  9605--9616, 2018.

\bibitem[Lutter et~al.(2019)Lutter, Ritter, and Peters]{lutter2019deep}
Lutter, M., Ritter, C., and Peters, J.
\newblock Deep lagrangian networks: Using physics as model prior for deep
  learning.
\newblock \emph{International Conference on Learning Representations}, 2019.

\bibitem[MacKay et~al.(2018)MacKay, Vicol, Ba, and
  Grosse]{mackay2018reversible}
MacKay, M., Vicol, P., Ba, J., and Grosse, R.~B.
\newblock Reversible recurrent neural networks.
\newblock In \emph{Advances in Neural Information Processing Systems}, pp.\
  9029--9040, 2018.

\bibitem[{Mnih} et~al.(2013){Mnih}, {Kavukcuoglu}, {Silver}, {Graves},
  {Antonoglou}, {Wierstra}, and {Riedmiller}]{mnih2013playing-atari-with}
{Mnih}, V., {Kavukcuoglu}, K., {Silver}, D., {Graves}, A., {Antonoglou}, I.,
  {Wierstra}, D., and {Riedmiller}, M.
\newblock {Playing Atari with Deep Reinforcement Learning}.
\newblock \emph{ArXiv e-prints}, December 2013.

\bibitem[Noether(1971)]{noether1971invariant}
Noether, E.
\newblock Invariant variation problems.
\newblock \emph{Transport Theory and Statistical Physics}, 1\penalty0
  (3):\penalty0 186--207, 1971.

\bibitem[Pukrittayakamee et~al.(2009)Pukrittayakamee, Malshe, Hagan, Raff,
  Narulkar, Bukkapatnum, and Komanduri]{pukrittayakamee2009simultaneous}
Pukrittayakamee, A., Malshe, M., Hagan, M., Raff, L., Narulkar, R.,
  Bukkapatnum, S., and Komanduri, R.
\newblock Simultaneous fitting of a potential-energy surface and its
  corresponding force fields using feedforward neural networks.
\newblock \emph{The Journal of chemical physics}, 130\penalty0 (13):\penalty0
  134101, 2009.

\bibitem[Reichl(1999)]{reichl1999modern}
Reichl, L.~E.
\newblock \emph{A modern course in statistical physics}.
\newblock AAPT, 1999.

\bibitem[Runge(1895)]{runge-1895-uber-die-numerische-auflosung}
Runge, C.
\newblock {\"U}ber die numerische aufl{\"o}sung von differentialgleichungen.
\newblock \emph{Mathematische Annalen}, 46\penalty0 (2):\penalty0 167--178,
  1895.

\bibitem[Rupp et~al.(2012)Rupp, Tkatchenko, M{\"u}ller, and
  Von~Lilienfeld]{rupp2012fast}
Rupp, M., Tkatchenko, A., M{\"u}ller, K.-R., and Von~Lilienfeld, O.~A.
\newblock Fast and accurate modeling of molecular atomization energies with
  machine learning.
\newblock \emph{Physical review letters}, 108\penalty0 (5):\penalty0 058301,
  2012.

\bibitem[Sakurai \& Commins(1995)Sakurai and Commins]{sakurai1995modern}
Sakurai, J.~J. and Commins, E.~D.
\newblock \emph{Modern quantum mechanics, revised edition}.
\newblock AAPT, 1995.

\bibitem[Salmon(1988)]{salmon1988hamiltonian}
Salmon, R.
\newblock Hamiltonian fluid mechanics.
\newblock \emph{Annual review of fluid mechanics}, 20\penalty0 (1):\penalty0
  225--256, 1988.

\bibitem[Santoro et~al.(2017)Santoro, Raposo, Barrett, Malinowski, Pascanu,
  Battaglia, and Lillicrap]{Santoro2017Simple}
Santoro, A., Raposo, D., Barrett, D.~G., Malinowski, M., Pascanu, R.,
  Battaglia, P., and Lillicrap, T.
\newblock A simple neural network module for relational reasoning.
\newblock In \emph{Advances in neural information processing systems}, pp.\
  4967--4976, 2017.

\bibitem[Schmidt \& Lipson(2009)Schmidt and Lipson]{Schmidt2009Distilling}
Schmidt, M. and Lipson, H.
\newblock Distilling free-form natural laws from experimental data.
\newblock \emph{Science}, 324\penalty0 (5923):\penalty0 81--85, 2009.

\bibitem[Sch{\"u}tt et~al.(2017)Sch{\"u}tt, Arbabzadah, Chmiela, M{\"u}ller,
  and Tkatchenko]{schutt2017quantum}
Sch{\"u}tt, K.~T., Arbabzadah, F., Chmiela, S., M{\"u}ller, K.~R., and
  Tkatchenko, A.
\newblock Quantum-chemical insights from deep tensor neural networks.
\newblock \emph{Nature communications}, 8:\penalty0 13890, 2017.

\bibitem[Silver et~al.(2017)Silver, Schrittwieser, Simonyan, Antonoglou, Huang,
  Guez, Hubert, Baker, Lai, Bolton, et~al.]{silver2017mastering}
Silver, D., Schrittwieser, J., Simonyan, K., Antonoglou, I., Huang, A., Guez,
  A., Hubert, T., Baker, L., Lai, M., Bolton, A., et~al.
\newblock Mastering the game of go without human knowledge.
\newblock \emph{Nature}, 550\penalty0 (7676):\penalty0 354, 2017.

\bibitem[Smith et~al.(2017)Smith, Isayev, and Roitberg]{smith2017ani}
Smith, J.~S., Isayev, O., and Roitberg, A.~E.
\newblock Ani-1: an extensible neural network potential with dft accuracy at
  force field computational cost.
\newblock \emph{Chemical science}, 8\penalty0 (4):\penalty0 3192--3203, 2017.

\bibitem[Taylor(2005)]{taylor2005classical}
Taylor, J.~R.
\newblock \emph{Classical mechanics}.
\newblock University Science Books, 2005.

\bibitem[Tenenbaum et~al.(2000)Tenenbaum, De~Silva, and
  Langford]{tenenbaum-2000-Science-a-global-geometric-framework}
Tenenbaum, J.~B., De~Silva, V., and Langford, J.~C.
\newblock A global geometric framework for nonlinear dimensionality reduction.
\newblock \emph{science}, 290\penalty0 (5500):\penalty0 2319--2323, 2000.

\bibitem[Tompson et~al.(2017)Tompson, Schlachter, Sprechmann, and
  Perlin]{tompson2017accelerating}
Tompson, J., Schlachter, K., Sprechmann, P., and Perlin, K.
\newblock Accelerating eulerian fluid simulation with convolutional networks.
\newblock In \emph{Proceedings of the 34th International Conference on Machine
  Learning-Volume 70}, pp.\  3424--3433. JMLR. org, 2017.

\bibitem[Wang et~al.(2018)Wang, Olsson, Wehmeyer, Perez, Charron, de~Fabritiis,
  Noe, and Clementi]{Wang2018Machine}
Wang, J., Olsson, S., Wehmeyer, C., Perez, A., Charron, N.~E., de~Fabritiis,
  G., Noe, F., and Clementi, C.
\newblock Machine learning of coarse-grained molecular dynamics force fields.
\newblock \emph{ACS Central Science}, 2018.

\bibitem[Watters et~al.(2017)Watters, Zoran, Weber, Battaglia, Pascanu, and
  Tacchetti]{Watters2017Visual}
Watters, N., Zoran, D., Weber, T., Battaglia, P., Pascanu, R., and Tacchetti,
  A.
\newblock Visual interaction networks: Learning a physics simulator from video.
\newblock In \emph{Advances in neural information processing systems}, pp.\
  4539--4547, 2017.

\bibitem[Yao et~al.(2018)Yao, Herr, Toth, Mckintyre, and
  Parkhill]{yao2018tensormol}
Yao, K., Herr, J.~E., Toth, D.~W., Mckintyre, R., and Parkhill, J.
\newblock The tensormol-0.1 model chemistry: A neural network augmented with
  long-range physics.
\newblock \emph{Chemical science}, 9\penalty0 (8):\penalty0 2261--2269, 2018.

\bibitem[Yosinski et~al.(2011)Yosinski, Clune, Hidalgo, Nguyen, Zagal, and
  Lipson]{yosinski2011evolving-robot-gaits}
Yosinski, J., Clune, J., Hidalgo, D., Nguyen, S., Zagal, J.~C., and Lipson, H.
\newblock Evolving robot gaits in hardware: the hyperneat generative encoding
  vs. parameter optimization.
\newblock In \emph{Proceedings of the 20th European Conference on Artificial
  Life}, pp.\  890--897, August 2011.

\end{thebibliography}
\bibliographystyle{hnn}

\appendix
\counterwithin{figure}{section}

\newpage

\section{Supplementary Information for Tasks 1-3} \label{appendix:learning}

\begin{figure}[H]
\begin{centering}
\subfigure[Baseline NN]{\figlabel{appendix:fig6a}\includegraphics[width=.36\columnwidth]{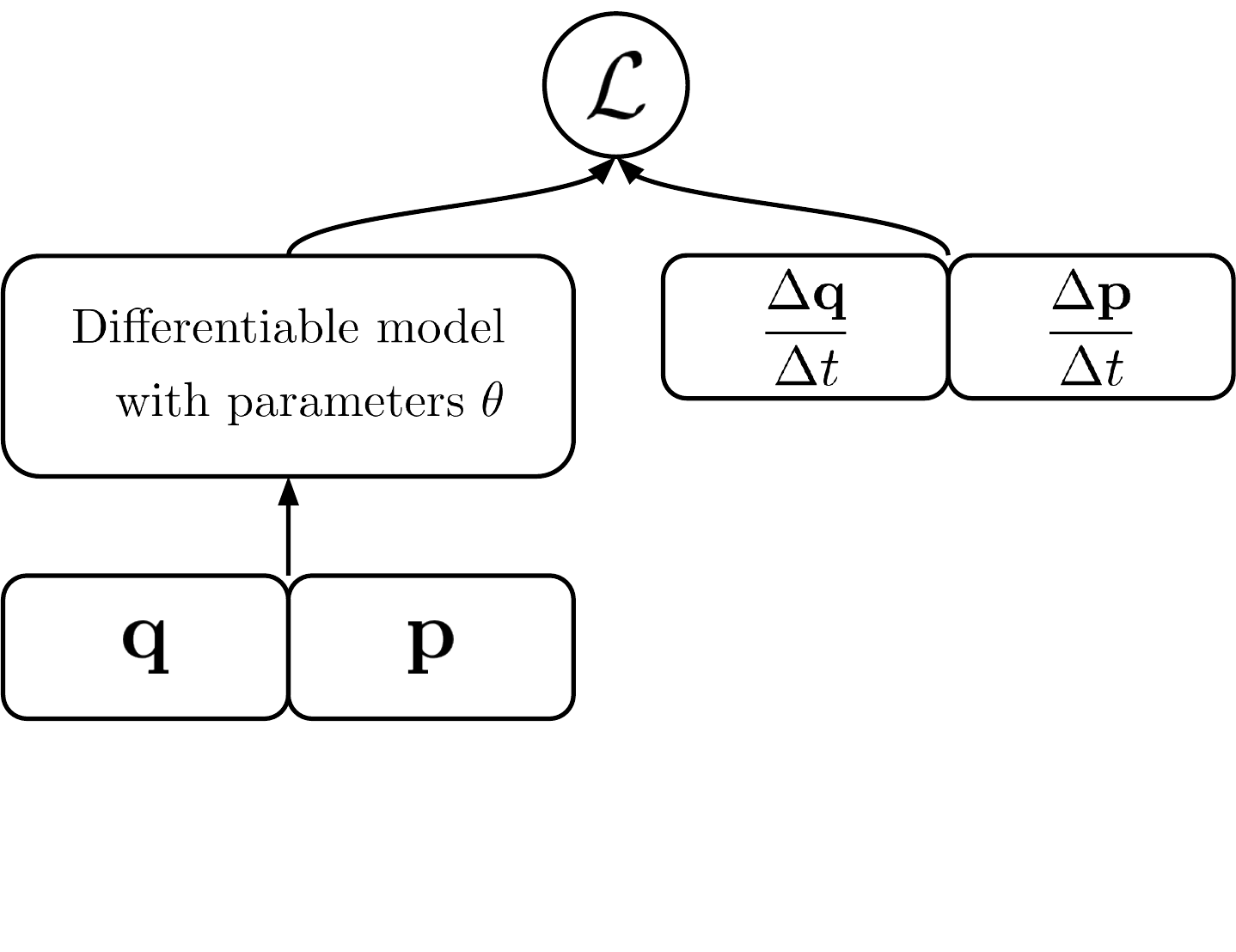}}
\hspace{1cm}
\subfigure[Hamiltonian NN]{\figlabel{appendix:fig6b}\includegraphics[width=.54\columnwidth]{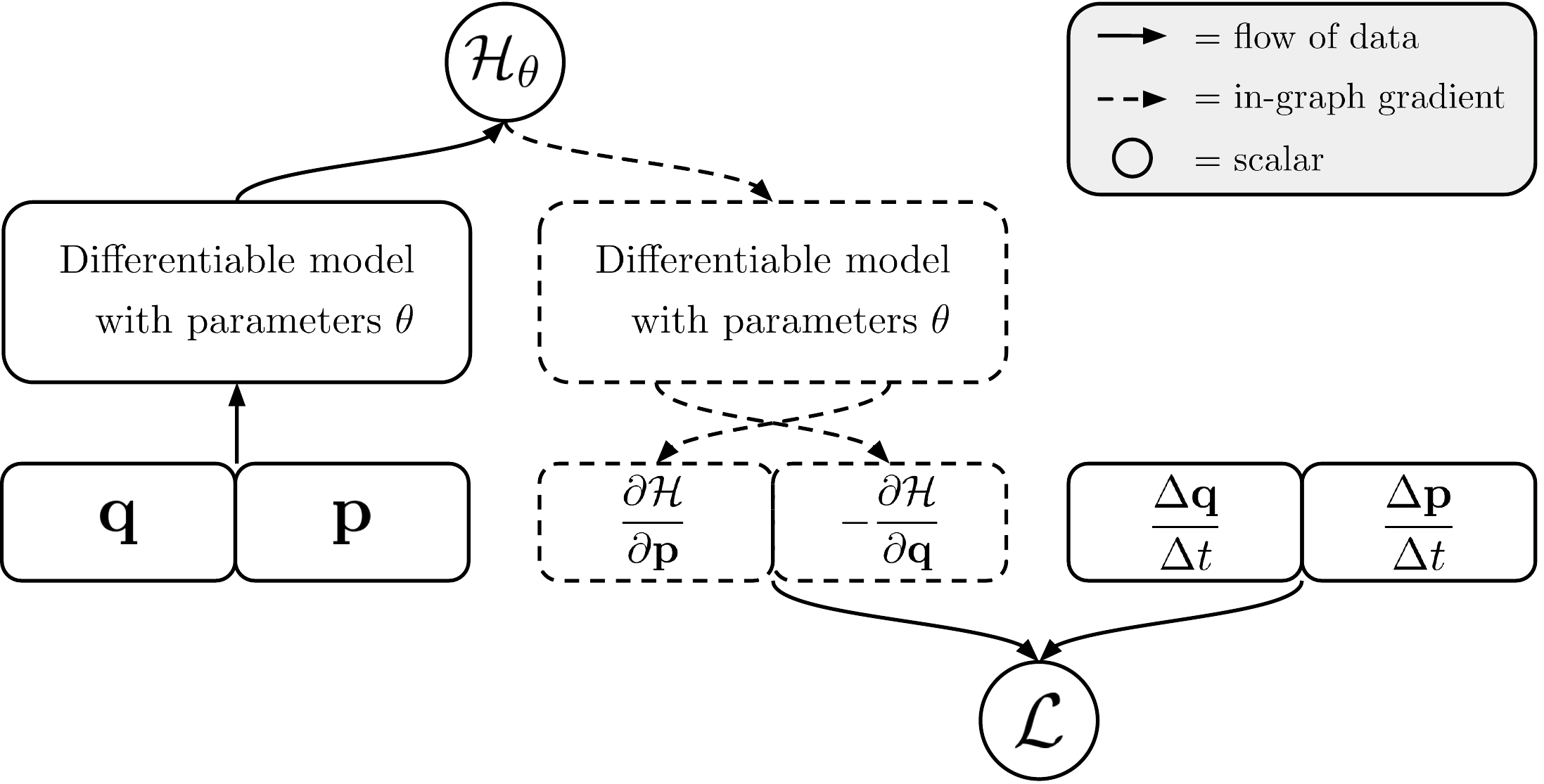}}
\end{centering}
\caption{HNN schema. The forward pass of an HNN is composed of a forward pass through a differentiable model as well as a backpropagation step through the model.}
\figlabel{appendix:fig6}
\end{figure}

\textbf{Training details.} We selected hyperparameters using a coarse grid search over learning rates $\{10^{-1}, 10^{-2}, 10^{-3}\}$, layer widths $\{100, 200, 300\}$, activations $\{\texttt{tanh}, \texttt{relu}\}$, and batch size where relevant $\{100, 200\}$. The main objective of this work was not to produce state-of-the-art results, so the settings we chose were aimed simply at producing models that gave good qualitative performance on the tasks at hand. We used weight decay of $10^{-4}$ on the first three tasks.

We trained all of these experiments on a desktop CPU.

\textbf{The train/test split on Task 3.} We partitioned the train and test sets on Task 3 in an unusual manner. The dataset provided by \cite{Schmidt2009Distilling} consisted of just a single trajectory from a real pendulum, as shown in the second panel of \figref{appendix:fig7c}. We needed to evaluate our model's performance over a series of adjacent time steps in order to measure the energy MSE metric. For this reason, we were forced to use the first $4/5$ of this trajectory for training (black vectors in \figref{appendix:fig7c}) and the last $1/5$ for evaluation (red vectors).

The consequence of this train/test split is that our test set had a slightly different distribution from our training set. We found that the relative magnitudes of the test losses between the baseline and HNN models were informative. We did not perform this ungainly train/test split on the other two tasks in this section.

\begin{figure}[H]
\centering
\subfigure[Task 1: Ideal mass-spring]{\figlabel{appendix:fig7a}\includegraphics[width=.8\textwidth]{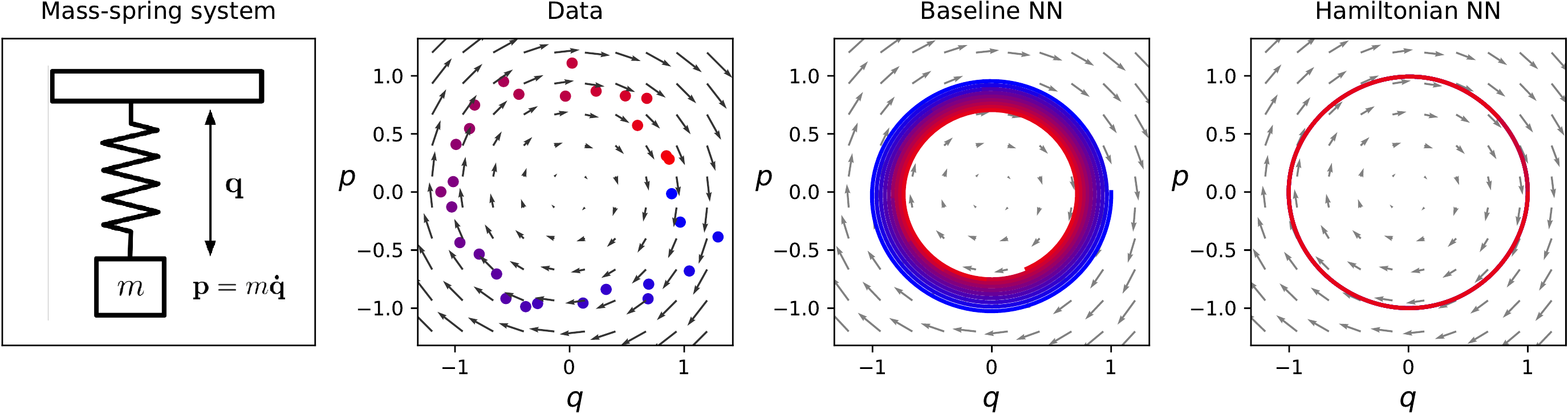}}
\subfigure[Task 2: Ideal pendulum]{\figlabel{appendix:fig7b}\includegraphics[width=.8\textwidth]{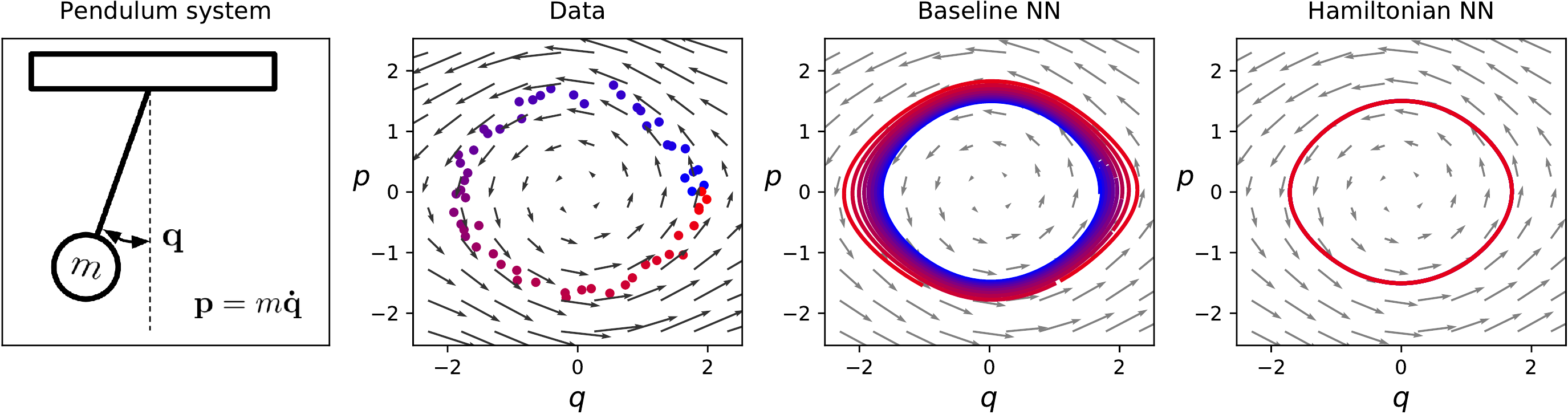}}
\subfigure[Task 3: Real pendulum]{\figlabel{appendix:fig7c}\includegraphics[width=.8\textwidth]{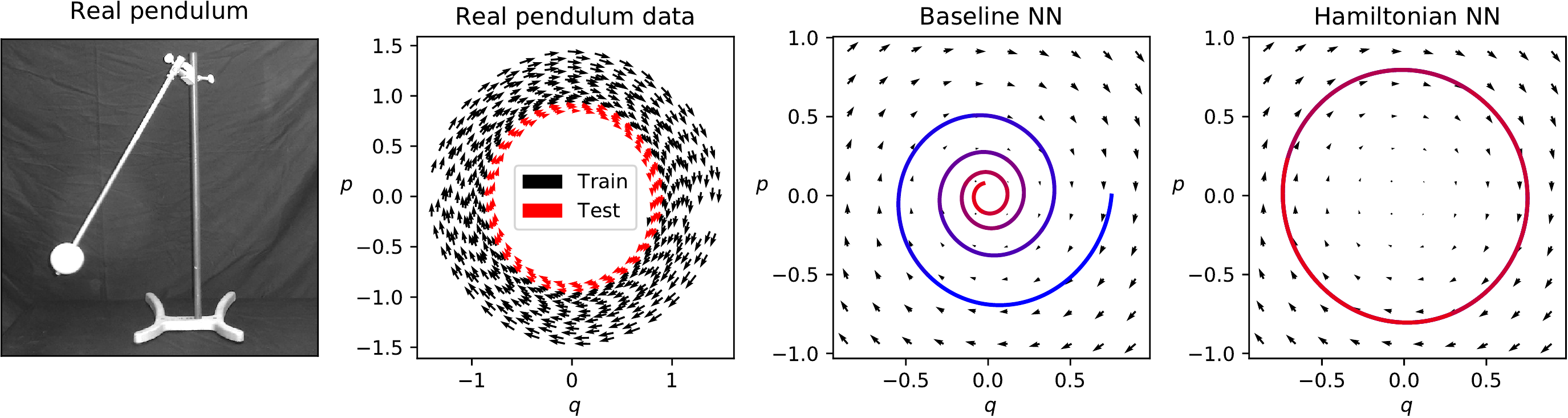}}
\caption{ More qualitative results comparing the HNN to a baseline neural network on the first three physics tasks. From top to bottom: Task 1: Ideal mass-spring, Task 2: ideal pendulum, Task 3: Real pendulum.}
\figlabel{appendix:fig7}
\end{figure}

\section{Supplementary Information for Task 4: Two-body problem} \label{appendix:orbits}

\textbf{Training details.} We selected hyperparameters with a grid search as described in the previous section. Again, the main objective of this work was not to produce state-of-the-art results, so the settings we chose were aimed simply at producing models that gave good qualitative performance on the tasks at hand. We did not use weight decay on this task, though when we tried a weight decay of $10^{-4}$ or results did not change significantly.

We trained this experiment on a desktop CPU.

\begin{figure}[H]
\centering
\subfigure[Baseline NN]{\figlabel{appendix:fig8a}\includegraphics[width=.8\textwidth]{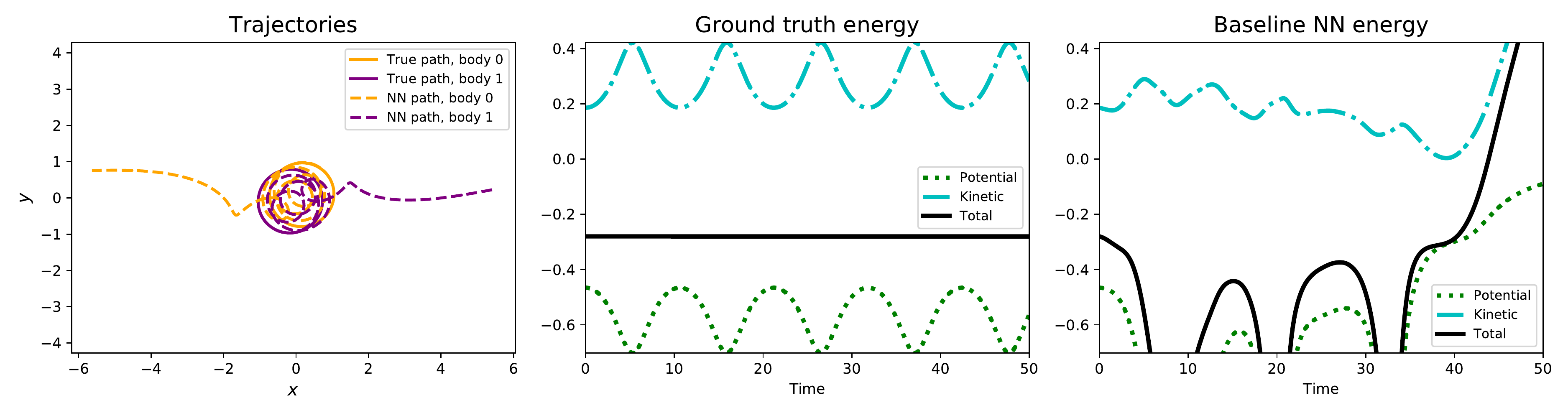}}\\
\subfigure[Hamiltonian NN]{\figlabel{appendix:fig8b}\includegraphics[width=.8\textwidth]{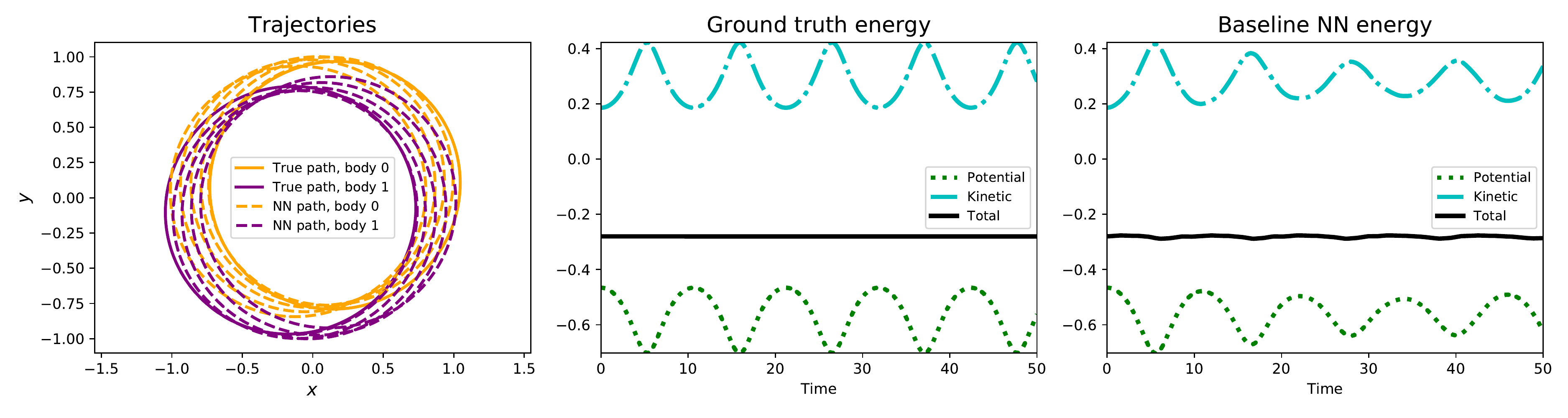}}
\caption{More qualitative results for the orbit task. Numerical errors accumulate in the baseline model until the bodies end up traveling in opposite directions. The total energy diverges towards infinity as well. In comparison, the HNN's trajectory diverges from the ground truth but continues to roughly conserve the total energy of the system.}
\figlabel{appendix:fig8}
\end{figure}

\begin{figure}[H]
\centering
\includegraphics[width=.95\textwidth]{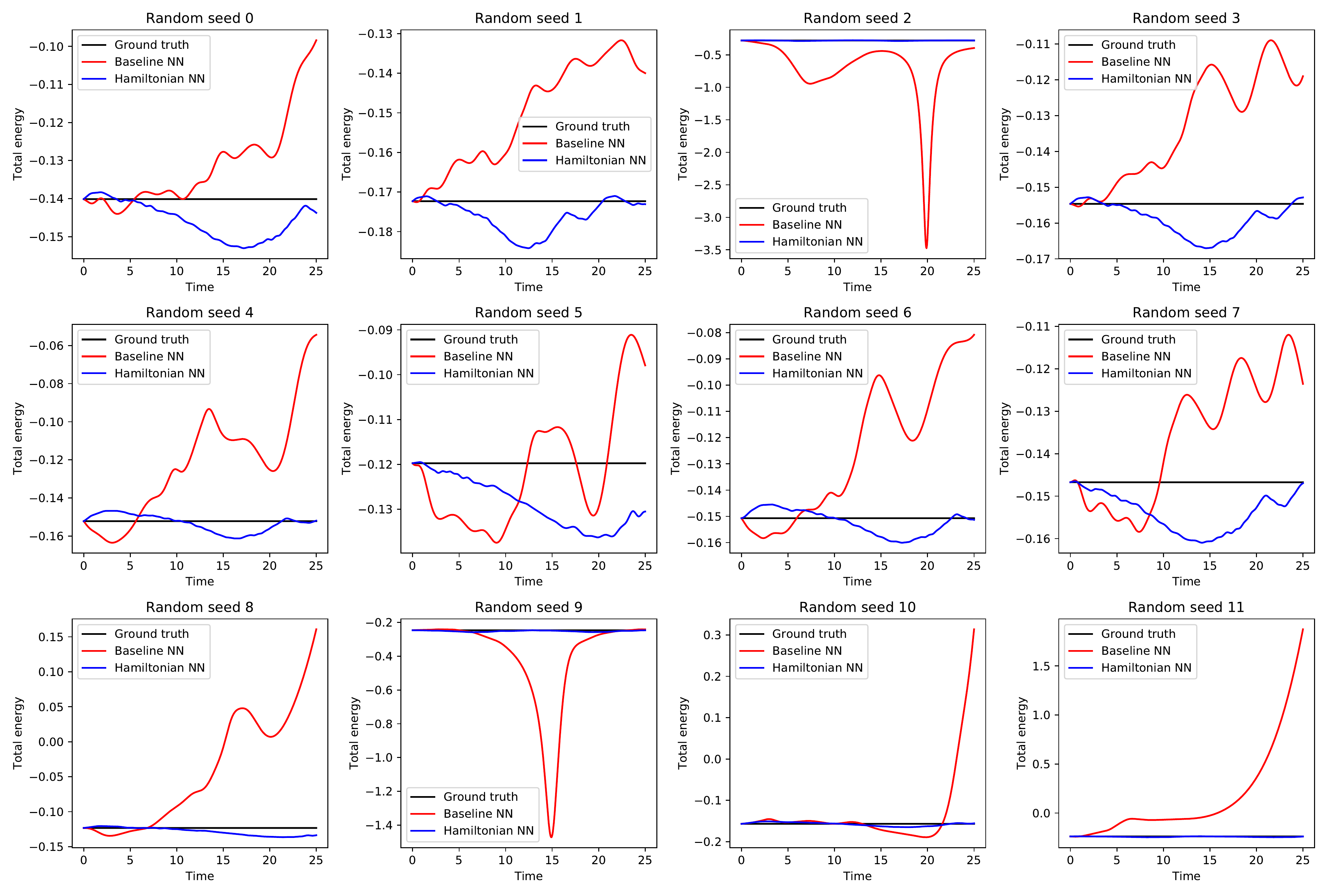}
\caption{Comparison of how well the HNN conserves total energy compared to the baseline its baseline on the two-body task.}
\figlabel{appendix:fig9}
\end{figure}

\textbf{Three body problem.} As mentioned briefly in the body of the paper, we also trained our models on the three body problem. The results we report here show a relative advantage to using the HNN over the baseline model. However, both models struggled to accurately model the dynamics of the three-body problem, which is why we relegated these results to the Appendix. Going forward, we hope to improve these results to the point where they can play a more substantial role in \secref{larger}.

\tabref{tab2} gives a summary of quantitative results and \figref{appendix:fig10} shows a qualitative analysis of the models we trained on this task.

\begin{table}[H]
\centering
\caption{Quantitative results for the three body problem. All values are multiplied by $10^{2}$. The confidence intervals suggest that both models struggle to model the distribution of the dataset. We hypothesize that the dynamic range of the dataset is a key issue.}
\tablabel{tab2}
\begin{tabular}{@{}lllllll@{}}
\toprule
& Train loss & & Test loss & & Energy MSE & \\ \midrule
Task & Baseline & HNN & Baseline & HNN & Baseline & HNN \\ \midrule
4b: Three body & $9.6\pm7$ & $8.0\pm2$ & $\mathbf{38\pm40}$ & $49\pm48$ & $\mathbf{1.1e4\pm8e3}$ & $4.2\pm3$  \\ \bottomrule
\end{tabular}
\end{table}

\begin{figure}[H]
\centering
\includegraphics[width=.8\textwidth]{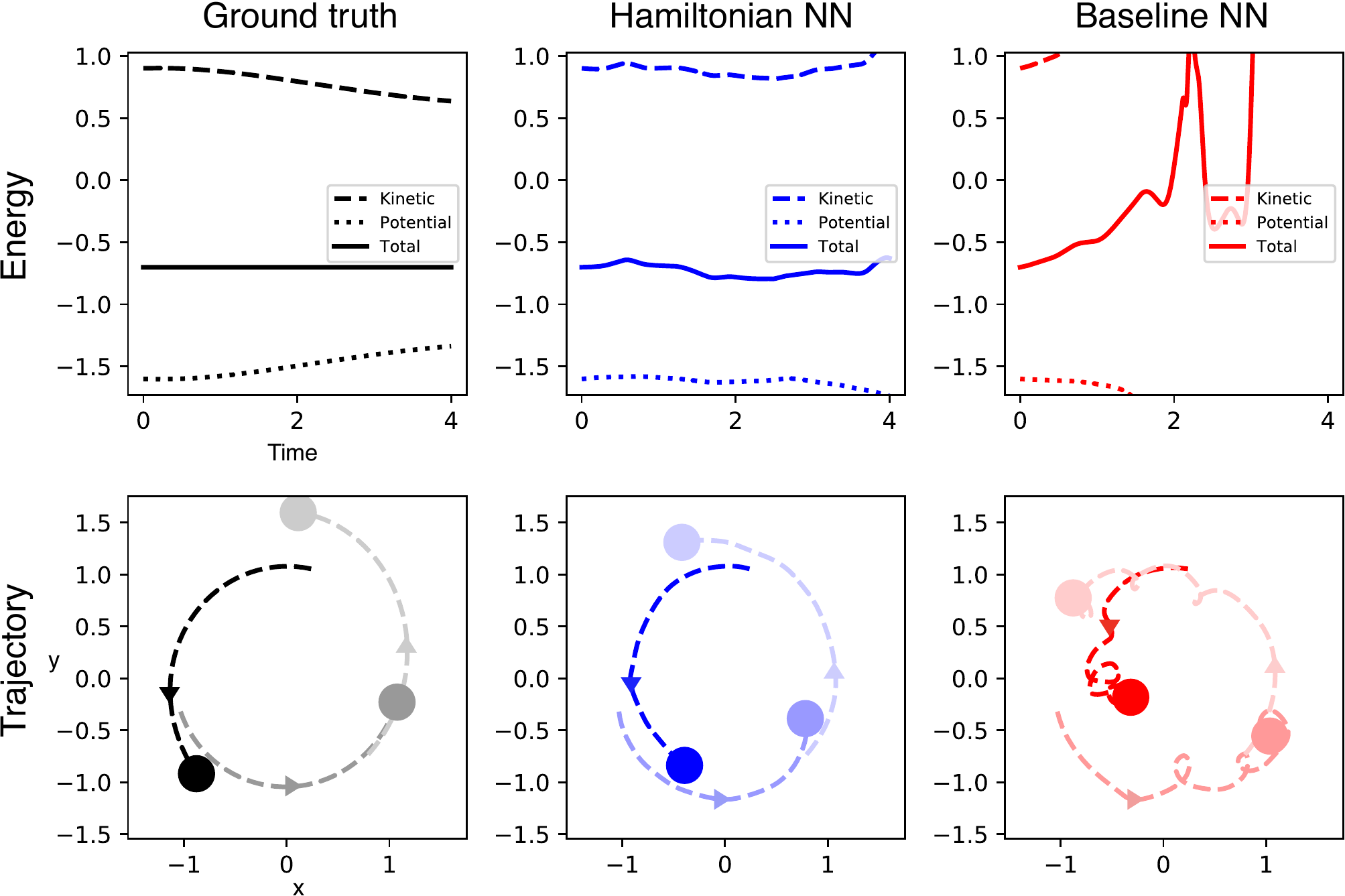}
\caption{
    Analysis of an example three-body trajectory. The baseline model does not conserve total energy and quickly diverges from ground truth. The HNN, meanwhile, roughly conserves total energy and its trajectories resemble the ground truth.
}
\figlabel{appendix:fig10}
\end{figure}

\section{Supplementary Information for Task 5: Pixel Pendulum} \label{appendix:pixels}

\textbf{Training details.} We selected hyperparameters with a grid search as described in the previous section. We used a weight decay of $10^{-5}$ on this experiment. We found that, unlike previous experiments, weight decay had a significant impact on results. We suspect that this is because the scale of the gradients on the weights of the HNN portion of the model were different from the scale of the gradients of the weights of the autoencoder portion of the model.

We trained this experiment on a desktop CPU.

\begin{figure}[H]
\centering
\subfigure[Latent space of the autoencoder]{\figlabel{appendix:fig11a}\includegraphics[width=.4\textwidth]{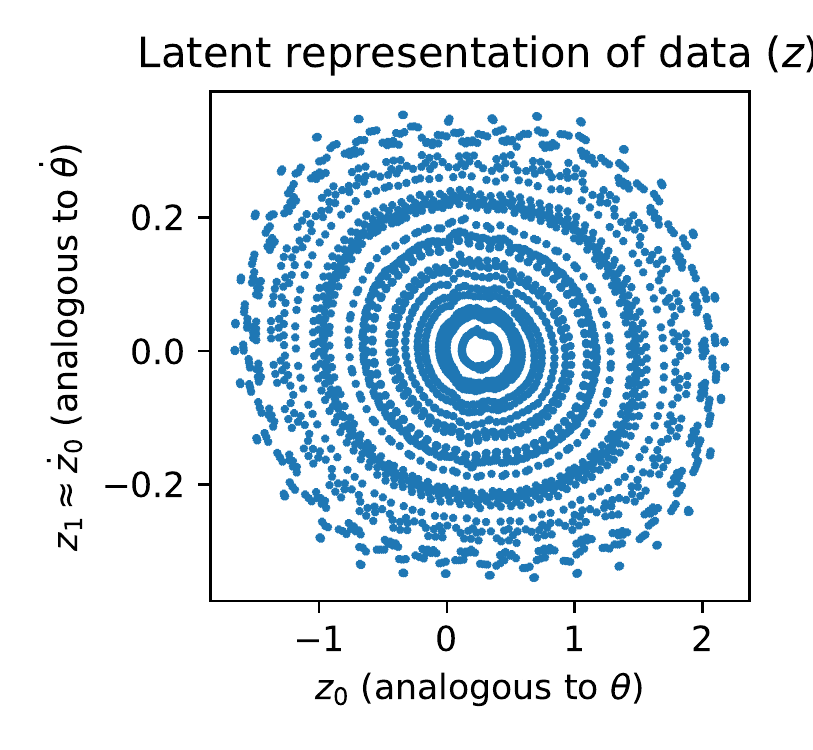}}
\subfigure[Contour plot of HNN-conserved quantity in latent space]{\figlabel{appendix:fig11b}\includegraphics[width=.4\textwidth]{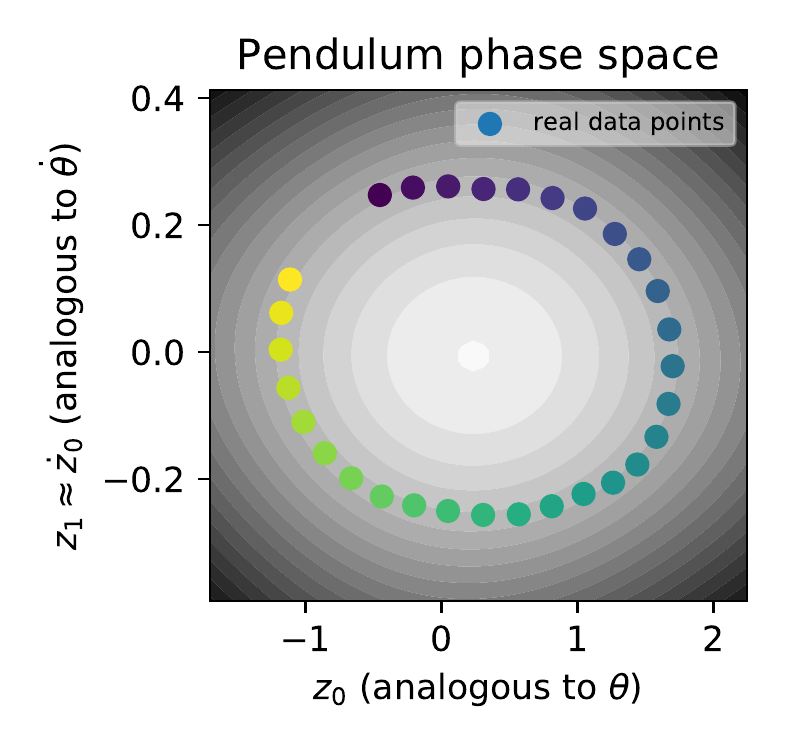}}
\caption{Latent space plots from the Pixel Pendulum model. Note that the learned latent space bears a strong resemblance to the true phase space of a pendulum. In particular, there is a faint diamond shape to the outer contour lines of \figref{appendix:fig11b}. This pattern is reminiscent of the nonlinear dynamics we observed in the ideal pendulum phase space plot of \figref{fig2} (row 2, column 1)}
\figlabel{appendix:fig11}
\end{figure}

\end{document}